\documentclass{article}

\newif\ifsubmit
\submittrue 

\usepackage{microtype}
\usepackage{graphicx}
\usepackage{subfigure}
\usepackage{booktabs}
\usepackage{hyperref}
\usepackage[accepted]{icml2018}
\usepackage{amsmath, amssymb, amsthm}
\usepackage{bm}
\usepackage{float}
\usepackage{url}

\ifsubmit
\newcommand{\todo}[1]{}
\newcommand{\mnote}[1]{}
\newcommand{\nnote}[1]{}
\newcommand{\inote}[1]{}
\newcommand{\onote}[1]{}
\newcommand{\enote}[1]{}
\else
\usepackage{xeCJK}
\newcommand{\todo}[1]{\textcolor{red}{TODO: #1}}
\newcommand{\mnote}[1]{\textcolor{blue}{(Miyato: #1)}}
\newcommand{\nnote}[1]{\textcolor{blue}{(Nakanishi: #1)}}
\newcommand{\inote}[1]{\textcolor{blue}{(Ichi: #1)}}
\newcommand{\onote}[1]{\textcolor{blue}{(Okanohara: #1)}}
\newcommand{\enote}[1]{\textcolor{brown}{(Editage: #1)}}
\fi

\newcommand{\enc}{F}
\newcommand{\dec}{G}
\newcommand{\aac}{H}

\newcommand{\qa}{Q}
\newcommand{\original}{\bm{x}}
\newcommand{\reconst}{\bm{\hat{x}}}
\newcommand{\precomp}{\bm{z}}
\newcommand{\comp}{\tilde{\bm{z}}}
\newcommand{\comps}{\bm{v}}
\newcommand{\save}{\bm{s}}
\newcommand{\bbR}{\mathbb{R}}

\icmltitlerunning{Neural Multi-scale Image Compression}
\usepackage{mathtools}

\begin{document}

\ifsubmit
\else
\begin{itemize}
    \item \todo{make citation style consistent}
    \item \todo{zの文字を入れ替えたので、すべての箇所（図も含めて)でconsistencyが取れているか確認}
    \item \todo{XeLaTeXからpdfLaTeXに変更する}
\end{itemize}
\fi

\twocolumn[
\icmltitle{Neural Multi-scale Image Compression}

\icmlsetsymbol{equal}{*}

\begin{icmlauthorlist}
\icmlauthor{Ken Nakanishi}{ut}
\icmlauthor{Shin-ichi Maeda}{pfn}
\icmlauthor{Takeru Miyato}{pfn}
\icmlauthor{Daisuke Okanohara}{pfn}
\end{icmlauthorlist}

\icmlaffiliation{ut}{The University of Tokyo}
\icmlaffiliation{pfn}{Preferred Networks, Inc}

\icmlcorrespondingauthor{Ken Nakanishi}{ken-nakanishi@g.ecc.u-tokyo.ac.jp}

\icmlkeywords{compression, deep learning, machine learning, image compression}

\vskip 0.3in
]

\printAffiliationsAndNotice{}

\begin{abstract}
This study presents a new lossy image compression method that utilizes the multi-scale features of natural images.
Our model consists of two networks:
 \textit{multi-scale lossy autoencoder} and \textit{parallel multi-scale lossless coder}.
The multi-scale lossy autoencoder extracts the multi-scale image features to quantized variables and the parallel multi-scale lossless coder enables rapid and accurate lossless coding of the quantized variables via encoding/decoding the variables in parallel.
Our proposed model achieves comparable performance to the state-of-the-art model on Kodak and RAISE-1k dataset images, and it encodes
a PNG image of size $768 \times 512$ in 70 ms with a single GPU and a single CPU process 
and decodes it into a high-fidelity image in approximately 200 ms.
\end{abstract}

\section{Introduction}
\begin{figure}[tbp]
    \centering
    \includegraphics[width=1.0\linewidth]{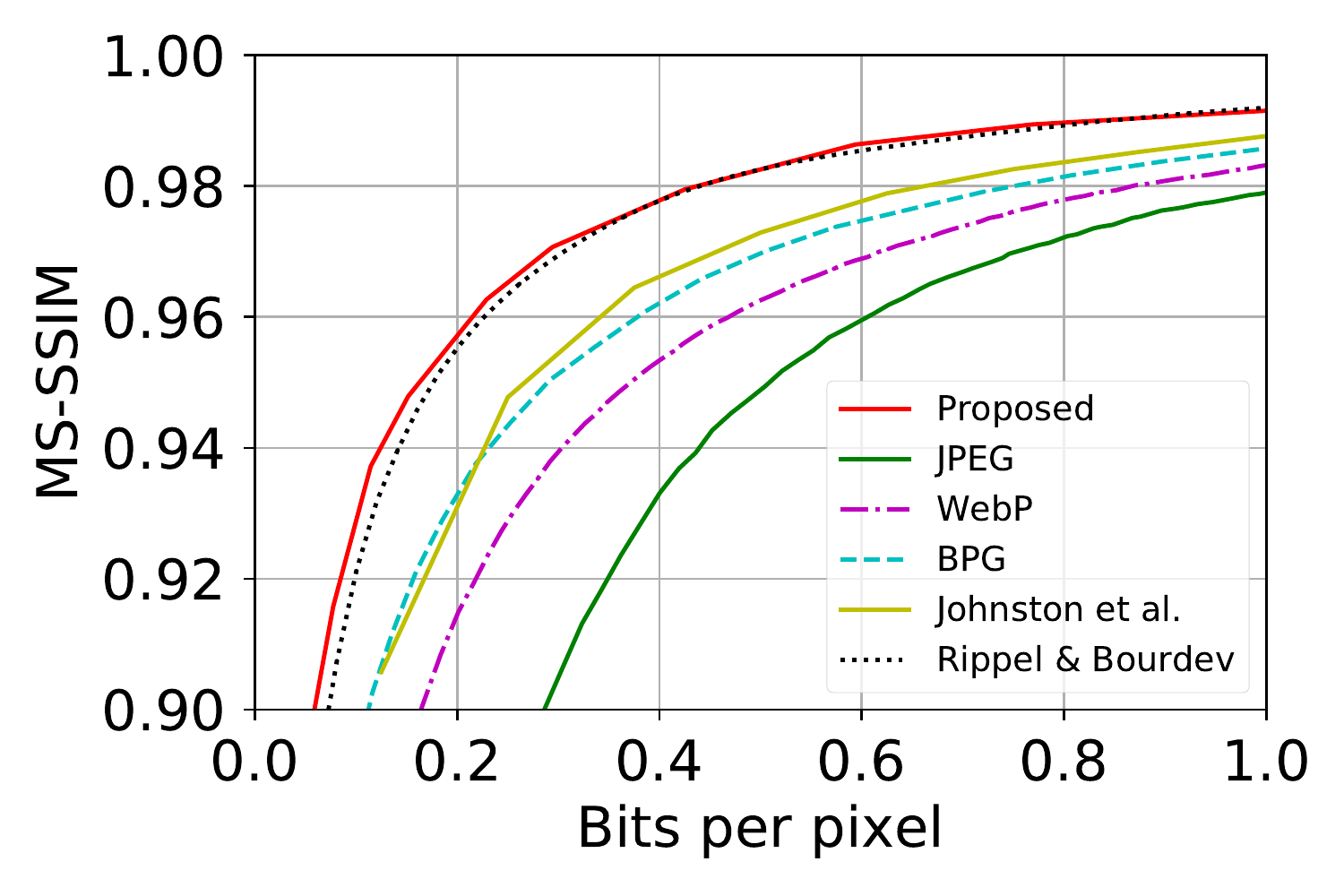}
    \caption{
        Rate-distortion trade off curves with different methods on Kodak dataset.
        The horizontal axis represents bits-per-pixel (bpp) and the vertical axis represents multi-scale structural similarity (MS-SSIM).
        Our model achieves better or comparable bpp with respect to the state-of-the-art results~\cite{rippel2017real}.
    }
    \label{fig:eval_kodak}
\end{figure}

\begin{figure}[tbp]
    \centering
    \includegraphics[width=0.8\linewidth]{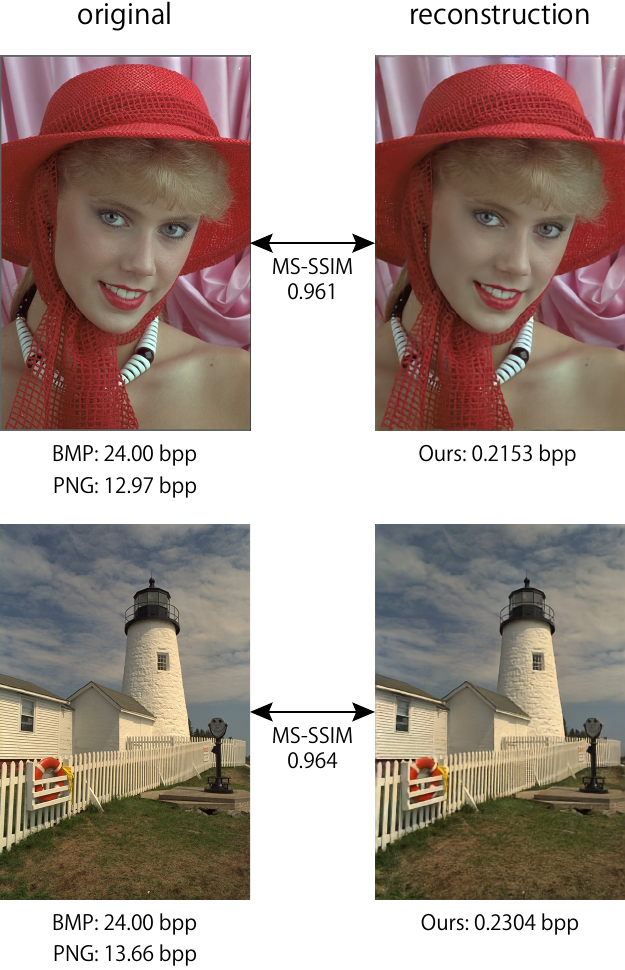}
    \caption{Original and reconstructed images by our model.}
    \label{fig:comp_reconst}
\end{figure}

\begin{figure}[tbp]
    \centering
    \includegraphics[width=1.0\linewidth]{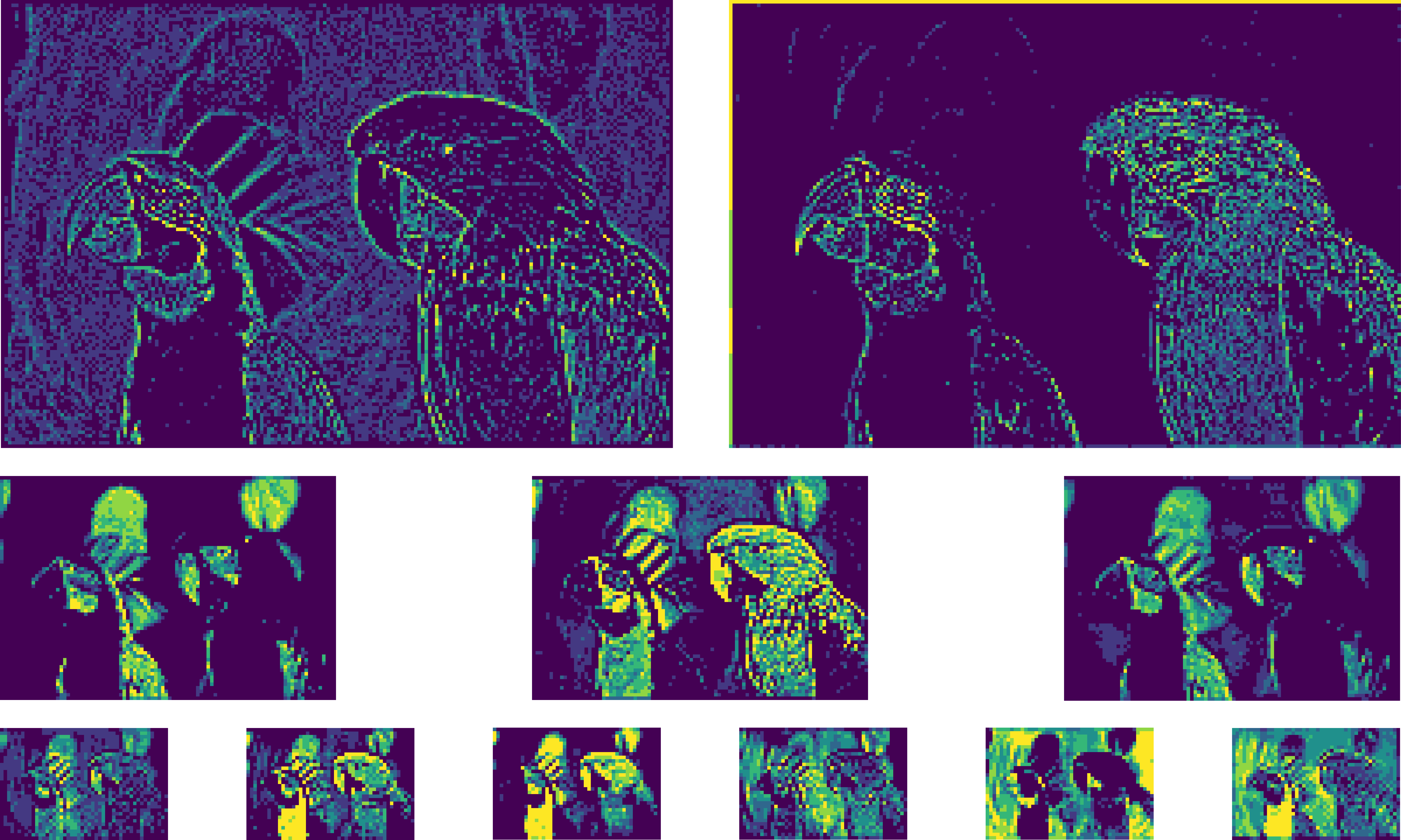}
    \caption{Visualization of quantized features, $\comp$, at each resolution layer. 
    Each panel represents the heatmap of $\comp$ at each resolution layer,
    where the top, middle, and bottom panels correspond to the map of the 1st, 2nd, and 3rd finest resolution layers.
    The value of $\comp$ in case of yellow-colored pixels is relatively high compared with the dark blue-colored pixels.
    }
    \label{fig:comp_image}
\end{figure}

Data compression for video and image data is a crucial technique for reducing communication traffic and saving data storage. 
Videos and images usually contain large redundancy,
enabling significant reductions in data size via \textit{lossy compression}, where data size is compressed while preserving the information necessary for its application.
In this work, we are concerned with lossy compression tasks for natural images.

JPEG has been widely used for lossy image compression.
However, the quality of the reconstructed images degrades, especially for low bit-rate compression. 
The degradation is considered to be caused by the use of linear transformation with an engineered basis. 
Linear transformations are insufficient for the accurate reconstruction of natural images,
 and an engineered basis may not be optimal.

In machine learning (ML)-based image compression, the compression model is optimized using training data.
The concept of optimizing the encoder and the decoder model via ML algorithm is not new. 
The $K$-means algorithm was used for vector quantization~\cite{Gersho12VQ},  
and the principal component analysis was used to construct the bases of transform coding~\cite{Goyal01PCA}. 
However, their representation power was still insufficient to surpass the performance of the engineered coders.
Recently, several studies proposed to use convolutional neural networks (CNN) for the lossy compression, 
resulting in impressive performance regarding lossy image compression~\cite{toderici2015variable,toderici2016full, balle2016end, theis2017lossy, Johnston2017improved, rippel2017real, mentzer2018conditional}
by exerting their strong representation power optimized via a large training dataset.

In this study, we utilize CNNs, but propose different architectures and training algorithm than those of existing studies to improve performance.
The performance targets are two-fold,
1. \textit{Good rate-distortion trade-off}
and 
2. \textit{Fast encoding and decoding}.
To improve these two points,
we propose a model that consists of two components: \textit{multi-scale lossy autoencoder} and \textit{parallel multi-scale lossless coder}. 
The former, multi-scale lossy autoencoder extracts the multi-scale structure of natural images via multi-scale coding to achieve better rate-distortion trade-off, while the latter, parallel multi-scale lossless coder facilitates the rapid encoding/decoding with minimal performance degradation.
We summarize the core concepts of each component of the model below. 
\begin{itemize}
    \item
    \textbf{Multi-scale lossy autoencoder}. 
    When we use a multi-layer CNN with pooling operation and/or strided convolution in this model, the deeper layers will obtain more global and high-level information from the image.
    Previous works \cite{toderici2015variable,toderici2016full, balle2016end, theis2017lossy, Johnston2017improved, rippel2017real, mentzer2018conditional} only used the features present at the deepest layer of such CNN model for encoding. 
In contrast, our lossy autoencoder model comprises of connections at different depths between the analyzer and the synthesizer, enabling encoding of multi-scale image features (See Fig.~\ref{fig:lossy_network}).   
    Using this architecture, we can achieve a high compression rate with precise localization.

    \item
    \textbf{Parallel multi-scale lossless coder}. 
Existing studies rely on sequential lossless coder, which makes the encoding/decoding time prohibitively large.
We consider concepts for parallel multi-scale computations based on the version of PixelCNN used in ~\cite{reed2017parallel} to enable
 encoding/decoding $\comp$ in a parallel manner; it achieves both fast encoding/decoding of $\comp$ and a high compression rate.

\end{itemize}
Our proposed model compresses Kodak\footnote{\url{http://r0k.us/graphics/kodak/}.} and RAISE-1k~\cite{Dang2015raise} dataset images into significantly smaller file sizes than JPEG, WebP, or BPG on fixed quality reconstructed images and achieves comparable rate distortion trade-off performance with respect to the state-of-the-art model~\cite{rippel2017real} on the Kodak and RAISE-1k datasets (See Figs.~\ref{fig:eval_kodak} and \ref{fig:eval_raise_crop}).
Simultaneously, the proposed method achieves reasonably fast encoding and decoding speeds. 
For example, our proposed model encodes
a PNG image of size $768 \times 512$ in 70 ms with a single GPU and a single CPU process 
and decodes it into a high-fidelity image with an MS-SSIM of 0.96 in approximately 200 ms.
Two examples of reconstruction images
 by our model with an MS-SSIM of approximately 0.96 are shown in Fig.~\ref{fig:comp_reconst}.

\section{Proposed method}
\subsection{Overview of proposed architecture}

\begin{figure*}[t]
    \centering
    \includegraphics[width=1.0\linewidth]{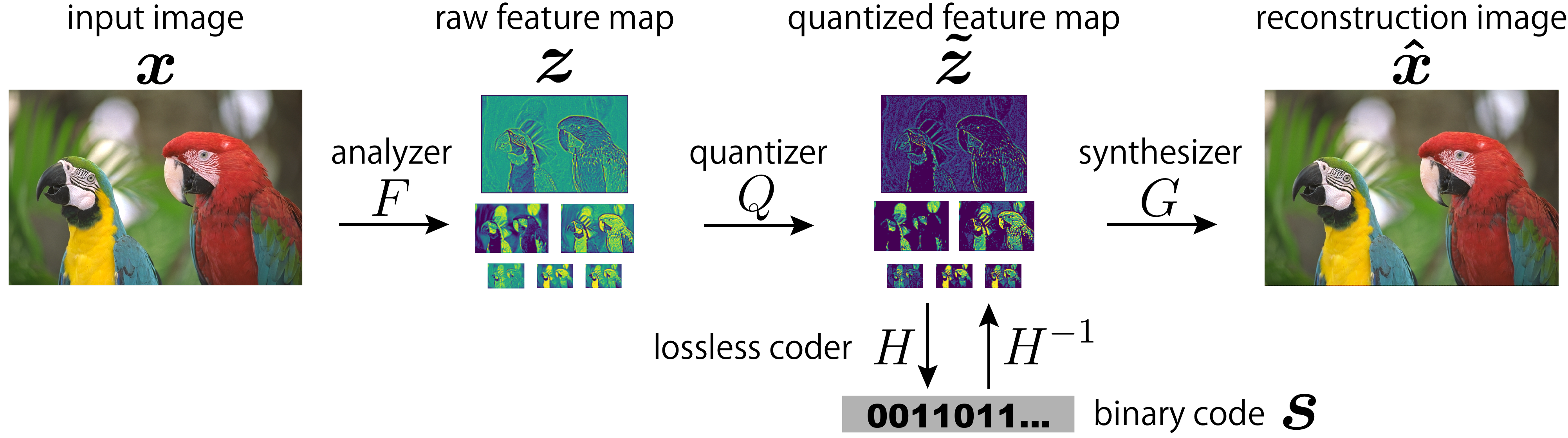}
    \caption{
        Overall architecture of proposed model. Our model consists of a lossy auto-encoder and a lossless coder where
        the lossy autoencoder consists of analyzer $\enc$, quantizer $\qa$, and synthesizer $\dec$. 
        The analyzer $\enc$ converts image $\original$ to a feature map $\precomp$.
        The quantizer module $\qa$ converts the feature map $\precomp$ to the quantized feature map $\comp$.
        The synthesizer $\dec$ converts the quantized feature map $\comp$ to a reconstruction image $\reconst$.
        The lossless coder $\aac$ compresses the quantized feature map $\comp$ to a binary code $\save$, losslessly.
        $\aac^{-1}$ is the inverse function of the lossless coder $\aac$.
    }
    \label{fig:overall_network}
\end{figure*}

In this section we first formulate the lossy compression. 
Subsequently, we introduce our lossy auto-encoder and lossless coder.
Let the original image be $\original$, the binary representation of the compressed variable be $\save$, and the reconstructed image from $\save$ be $\reconst$.
The objective of the lossy image compression is to 
minimize the code length (i.e., file size) of $\save$ while minimizing the distortion  between $\original$ and $\reconst$: $d(\original, \reconst)$ as much as possible. 
The selection of the distortion
$d(\cdot, \cdot)$ is arbitrary as long as it allows differentiation 
with respect to the input image. 

Our model consists of a lossy auto-encoder and a lossless coder as shown in Fig.~\ref{fig:overall_network}.
The auto-encoder transforms the original image $\original$ into the features $\precomp$ 
using the analyzer~$\enc$ as $\precomp = \enc(\original;\phi)$.
Subsequently, the features are quantized using a quantizer $\qa$ as $\comp = \qa(\precomp)$.
$\qa$ quantizes each element of $\comp$ using a multi-level uniform quantizer, which has no learned parameters.
Finally, the synthesizer $\dec$ of the auto-encoder, recovers the reconstructed image, $\reconst = \dec(\comp;\theta)$.
Here, $\phi$ and $\theta$ are the parameters of $\enc$ and $\dec$, respectively.
Parameters $\theta$ and $\phi$ are optimized to minimize the following distortion loss:
\begin{align}
    L(\theta, \phi) \coloneqq \mathbb{E}_{p(\original)} \left[d(\original, \dec(\qa(\enc(\original;\phi));\theta))\right],  \label{eq:lossy_loss}
\end{align}
where $\mathbb{E}_{p(\original)} [\cdot]$ represents the expectation over the input distribution, which is approximated by an empirical distribution.

The second neural network is used for the lossless compression of the quantized features $\comp$.
According to Shannon's information theory,
the average code length is minimized when we allocate the code length, $\log_2 p(\comp)$ bits,
for the signal $\comp$ whose occurrence probability is $p(\comp)$.
Hence, we estimate the occurrence probability of $\comp$ as $p(\comp; \omega)$ where $\omega$ is a parameter to be estimated.
In this study, $\omega$ is estimated via maximum likelihood estimation. 
Thus, we minimize cross entropy between $p(\comp)$ and $p(\comp; \omega)$,
 using the fixed analyzer and synthesizer:
\begin{align}
H(p, p_{\omega}) \coloneqq -\mathbb{E}_{p(\comp)}[\log_2 p(\comp; \omega)]. \label{eq:lossless_loss}
\end{align}

Note that our objective function for $\phi, \theta$, and $\omega$ can be easily extended to the rate-distortion  
cost function, a weighted sum of distortion loss~\eqref{eq:lossy_loss}, and cross entropy~\eqref{eq:lossless_loss}.
Using the rate-distortion cost function, we can jointly optimize $\phi, \theta$, and $\omega$, as in previous studies~\cite{balle2016end, theis2017lossy, mentzer2018conditional}.
Nevertheless, we separately optimize them by first optimizing the distortion loss with respect to $\phi$ and $\theta$. Subsequently, we optimize the cross entropy with respect to $\omega$.
This two-step optimization simplifies the optimization of the analyzer $\enc$.
Because the derivative of the rate-distortion function with respect to the parameter of the analyzer, $\phi$ depends on the occurrence probability $p(\comp; \omega)$,
the computation of the derivative requires time and becomes complex. 
Furthermore, it may consume excessive memory for the computation of
 the derivative when we optimize the parameters with large number of images.
Optimization with the rate-distortion cost function could be 
a future direction of research to pursue further performance improvements.

\subsection{Multi-scale Auto-encoder} \label{sec:lossy}

\begin{figure*}[t]
    \centering
    \includegraphics[width=1.0\linewidth]{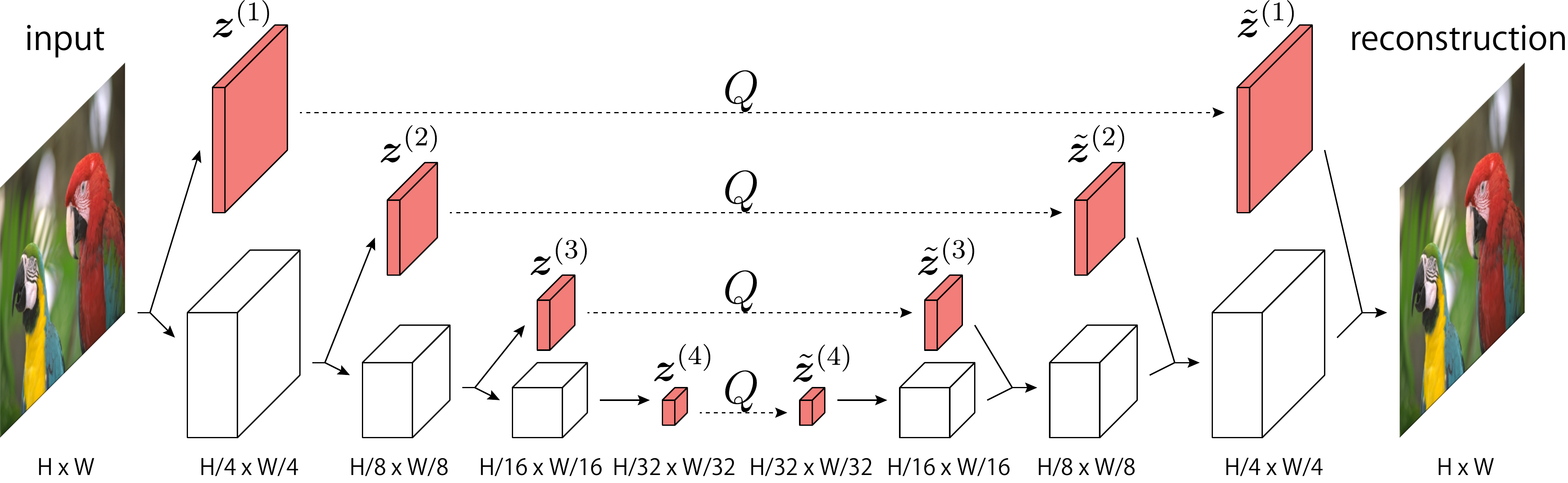}
    \caption{
        Architecture of our multi-scale  autoencoder.
        H and W are height and width of an input image, respectively.
        The left side of the network is the analyzer $\enc$, and the right side is the synthesizer $\dec$.
        Light and vivid red-colored boxes represent pre-quantized variables $\precomp$, and quantized variables $\comp$.
        The analyzer $\enc$ and synthesizer $\dec$ are described in Section \ref{sec:lossy}.
        Quantizer module $\qa$ is described in Section \ref{sec:qa}.
    }
    \label{fig:lossy_network}
\end{figure*}

In this section, we describe our proposed auto-encoder.
Our multi-scale auto-encoder consists of an analyzer $\enc$, a synthesizer $\dec$, and a quantizer $\qa$.
Both the analyzer $\enc$ and synthesizer $\dec$ are composed of CNNs similar to the existing studies.
The difference between the proposed and existing models is that 
our auto-encoder encodes information of the original image in multi-scale features, as follows.
\begin{align}
   \enc &: \original  \mapsto \{\precomp^{(i)}\}_{i=1}^{M}, \\
      \qa &: \{\precomp^{(i)}\}_{i=1}^{M} \mapsto \{\comp^{(i)}\}_{i=1}^{M} \\
   \dec &:  \{\comp^{(i)}\}_{i=1}^{M} \mapsto \reconst, \\
   \precomp^{(i)} &\in \bbR^{C^{(i)} \times H^{(i)} \times W^{(i)}}, \\
   \comp^{(i)} &\in \{0,\dots, N-1\}^{C^{(i)} \times H^{(i)} \times W^{(i)}}, 
\end{align}
where $\precomp^{(i)}$ and $\comp^{(i)}$ denote the
 $i$-th ($i=1, \cdots, M$) resolution of features and its quantized version, 
 whose spatial resolution is $H^{(i)} \times W^{(i)}$ and  number of channels is $C^{(i)}$.
The spatial resolution becomes coarser as the layer $i$ becomes deeper, such that both
$H^{(i)} > H^{(i+1)}$ and $W^{(i)} > W^{(i+1)}$ hold.
Each element of $\comp$ is quantized into $\{0,\dots,N-1\},~N \in \mathbb{N}$.
Fig.~\ref{fig:lossy_network} shows an example of $M=4$.
Global and coarse information including textures, are encoded at the deeper layer, whereas local and fine information, such as edges, are encoded at the shallower layer. 
The parameters of $\enc$ and $\dec$, $\phi$ and $\theta$, are trained to minimize distortion loss \eqref{eq:lossy_loss}. 

\subsubsection{Quantizer module} \label{sec:qa}

Because quantization is not a differentiable operation,
it makes optimization difficult.
Recent studies \cite{balle2016end, toderici2016full, Johnston2017improved, Agustsson2018softhard} 
have used stochastic perturbations to avoid the problem of non-differentiability. 
They replace the original distortion loss with the 
average distortion loss where the distortion occurs owing to the 
injection of the stochastic perturbation to the feature maps, 
instead of the deterministic quantization.
In general, however, the stochastic perturbation makes the training longer,
because we require considering samples with large size to approximate the expected value . 

\begin{figure}[t]
    \centering
    \includegraphics[width=\linewidth]{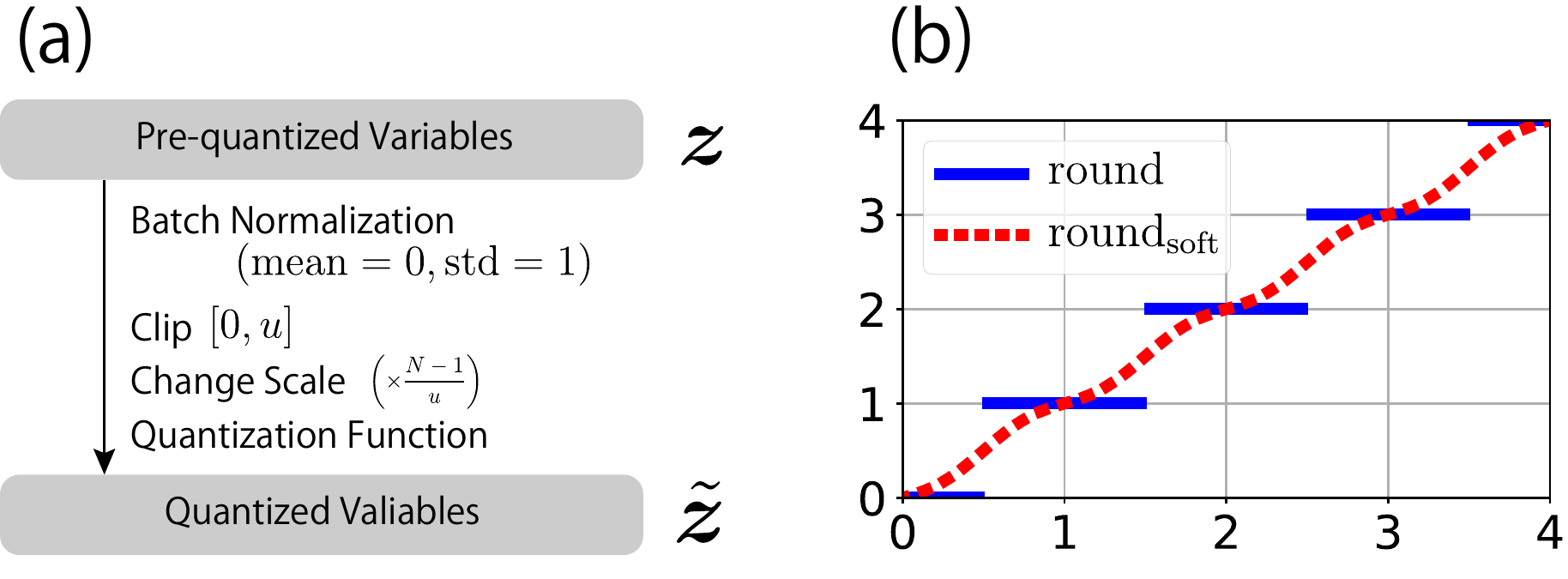}
    \caption{Quantizer module.
    Panel (a) shows quantization procedures and  
    panel (b) shows ${\rm round}(x)$ (blue dashed line) and the plot of ${\rm round}_{\rm soft}(x)$ (red solid line), which are used as a hard quantization function and a soft quantization function, respectively. 
}
    \label{fig:quantize_network}
\end{figure}

To avoid complexity of optimization when injecting stochastic perturbation,
we adopt deterministic quantization even during training,  
similar to the recent work~\cite{mentzer2018conditional}.
The quantizer module we use is shown in Fig.~\ref{fig:quantize_network}.

Before quantization, we apply several preprocessing steps on $\precomp$.
First, to enhance the compression rate of the lossless coding,
we apply batch normalization (BN) on $\precomp$, which 
drives the statistics of the feature map to exhibit zero mean and unit variance.
BN makes each channel of $\precomp$ possess similar statistics to each other,
which is beneficial for estimating the probability of the discretized feature map $p(\comp; \omega)$ using which CNN shares internal layer except for the top layer.
Additionally, the control of statistics via BN is advantageous 
for decreasing the entropy of $\comp$. 
Subsequently, we clip the normalized $\precomp$ into $[0, u]$        $(u>0)$
and expand its range into $[0,N-1]$ via multiplying $\frac{N-1}{u}$.
In this study, we set $u=4$.

Following the above mentioned preprocessing, we apply multi-level quantization:
\begin{equation}
{\rm round}(x) = \lceil x-0.5 \rceil,  \label{eq:hardq}
\end{equation}
where $\lceil a \rceil$
is a ceiling function that yields the smallest integer, which is larger than or equal to $a$.
This quantization function is not differentiable and 
does not allow conduction of the gradient-based optimization.
To overcome this difficulty, we consider a similar strategy as~\citet{mentzer2018conditional}. 
We replace the quantization function with the ``soft" quantization function 
when computing back-propagation, while the intact quantization function is used 
to compute forward propagation.
The soft quantization function we used is written as  
\begin{align}
   {\rm round}_{\rm soft}(x) = x - \alpha \frac{\sin(2\pi x)}{2\pi} \label{eq:softq},
\end{align}
where we used $\alpha=\frac{1}{2}$. 
Fig.~\ref{fig:quantize_network} (b) shows ${\rm round}_{\rm soft}(x)$ 
(Eq.~\eqref{eq:softq}) overlaid on ${\rm round}(x)$~(Eq.~\eqref{eq:hardq}).
Owing to this approximation, we can conduct conventional gradient-based optimization 
for the minimization of the distortion loss (Eq.~\eqref{eq:lossy_loss}) with respect to $\theta$ and $\phi$ and the usual training of the neural network. 
Although this gradient-based optimization uses improper gradient, 
the performance of our model is comparable or superior 
to the performance of existing models, 
as demonstrated in our experiment, implying that the side effect is not prominent.

\subsection{Parallel Multi-scale Lossless Coder}\label{sec:lossless}
In this section, we explain the construction of
lossless coding , $\aac \colon \comp \mapsto \save$, 
 that transmits the multi-scale feature map $\comp$ 
 into a one-dimensional binary sequence $\save$.

\begin{figure}[t]
    \centering
    \includegraphics[width=0.9\linewidth]{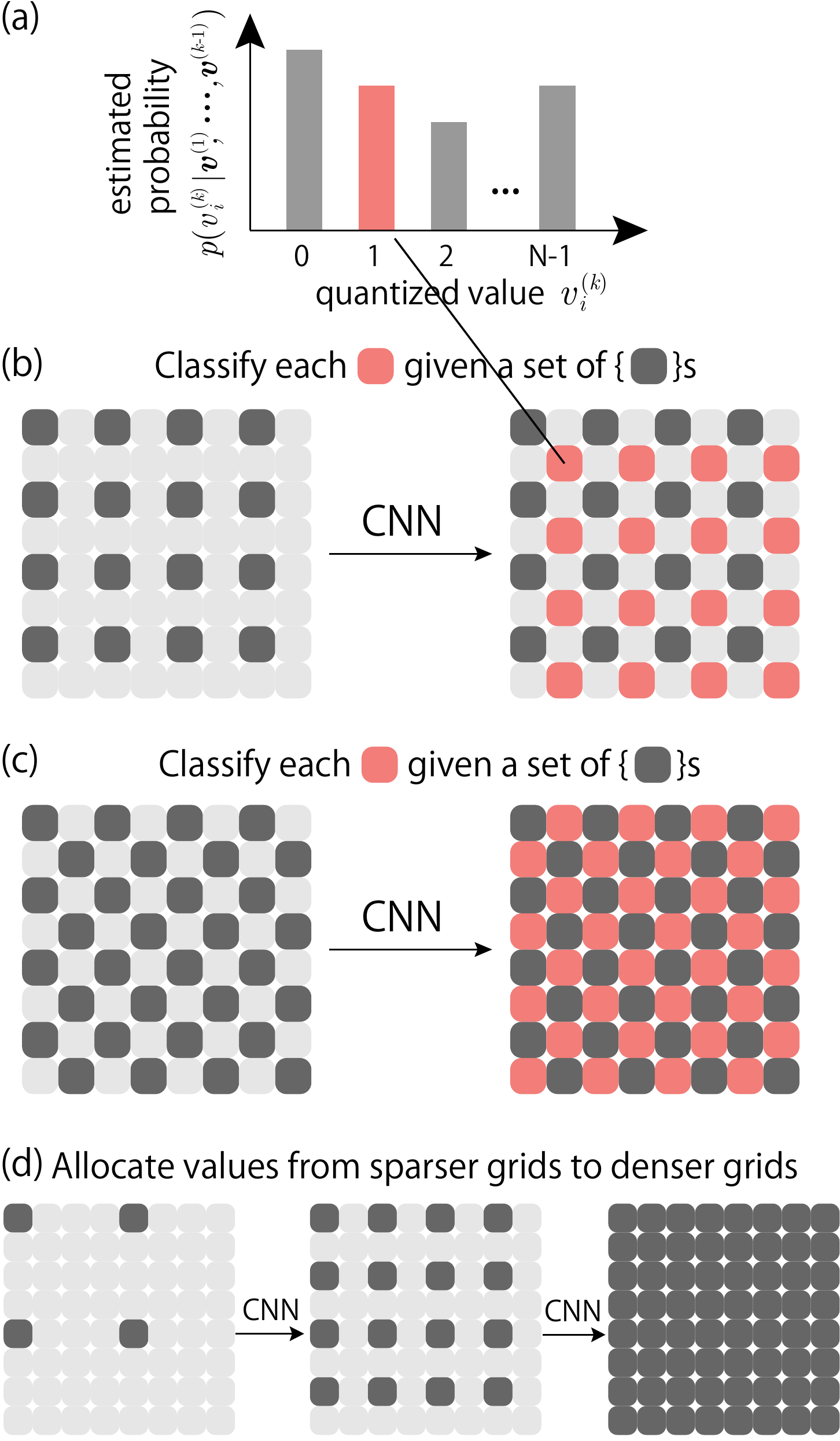}
    \caption{Lossless coding with parallel multi-scale PixelCNN.
        (a) The conditional probability estimated by CNN is used to 
        encode the $i$-th quantized feature $\comps^{(i)}$; 
        (b),(c) two types of conditional independence in spatial domain are assumed, and each distribution is estimated by solving the classification task with CNN; and 
        (d) the values of dense grids are encoded by conducting procedures (b) and (c), iteratively.
    }
    \label{fig:lossless_unit}
\end{figure}

To minimize average code length, we estimate the occurrence probability of $\comp$.
Suppose $\comp$ is indexed in raster scan order as $\comp = (z_1, \cdots, z_I)$ where $I=\sum_{i=1}^M C^{(i)}\times H^{(i)}\times W^{(i)}$.
Subsequently, the joint probability $p(\comp)$ is represented as the product of the conditional distributions: 
\begin{equation}
p(\comp) = p(z_1)\prod_{i=2}^I p(z_i | z_1, \cdots, z_{i-1}).
\end{equation}
Conveniently, the problem of learning the conditional distribution can be formulated
as a supervised classification task, which is successfully solved via neural networks.
If $z_i$ takes one of $N$-values, the neural network that contains $N$-output variables is trained to 
predict the value (i.e., label) of $z_i$.
Subsequently, the output of the trained neural network mimics $p(z_i | z_1, z_2, \cdots, z_{i-1})$. 
The estimated probability is used for encoding $z_i$, 
as illustrated in  Fig.~\ref{fig:lossless_unit}(a).

\citet{toderici2016full} used PixelRNN to learn 
$p(z_i | z_1,  \cdots, z_{i-1})$ ($i \geq 2$),  
demonstrating that it achieves a high theoretical compression rate.
In practice, however, it requires a long computational time for both encoding and decoding,  
proportional to the number of elements of $\comp$, 
because PixelRNN sequentially encodes $z_i$.

To reduce the computation time for both encoding and decoding, 
we use a parallel multi-scale PixelCNN~\cite{reed2017parallel}.
The concept behind this model is to take advantage of conditional independence. 
We divide the elements of $\comp$ into $K$ subsets $\comps^{(k)} \quad (k = 1,2,\cdots, K)$, 
where the $k$-th subset includes $I_k$ elements. 
We subsequently assume the conditional independence among the elements of the subset.
Namely, we assume that $p(\comp)$ is represented as 
\begin{equation} \label{eq:subset}
p(\comp) \simeq p(\comps^{(1)})\prod_{k=2}^K \prod_{i=1}^{I_k} p(v^{(k)}_i | \comps^{(1)}, \cdots, \comps^{(k-1)}).
\end{equation}
Although conditional independence does not hold in general,
it significantly reduces the computation time, 
because the number of evaluations of neural networks is 
no longer proportional to the number of elements $N$, but 
is instead proportional to the number of subsets $K$,  
where, typically, $K \ll N$.

Specifically, we assume conditional independence in spatial and channel domain as conducted in \citet{reed2017parallel}, but with a slightly different implementation.
Because we use CNN as the analyzer and synthesizer of the autoencoder,
the feature map $\comp$ preserves the spatial information. 
The spatial correlation between pixels of $\comp$ tends to decrease as the distance increases.
Thus, we assume the conditional independence between the distant units in the feature map when
conditioned on the relatively close units in the feature map. 
This is illustrated in Figs.~\ref{fig:lossless_unit}(b)-(d), 
where the red units are assumed to be independent from each other under the condition 
that the dark-gray units are provided.
We simply consider a single resolution case; the multi-resolution case is explained in appendix.
Encoding the red units given the dark-gray 
units, as in the order of Fig.~\ref{fig:lossless_unit} (b) and  (c),
we can encode denser units, based on the given sparser units.
Iterating this procedure, as shown in Fig.~\ref{fig:lossless_unit} (d), we can encode all the units of $\comp$, given $\comps^{(1)}$.
To encode $\comps^{(1)}$
, we simply assume independence among the units in $\comps^{(1)}$
and assume it obeys an identical distribution, irrelevant to the spatial position. 
Subsequently, we estimate the histogram to approximate the distribution.

\section{Comparison with existing CNN-based lossy image compression} \label{sec:related_work}
In this section, we review recent studies regarding CNN-based image compression to elucidate the value of our contribution. 
CNN-based image compression exhibits high image compression performance.
It also achieves better visual quality of reconstruction, especially on low bit-rate compression, compared with classic compression methods, such as JPEG.

For the entropy coding, \citet{balle2016end} and \citet{theis2017lossy} modeled $p(\comp)$ using a factor of independent probability distributions (i.e., $p(\comp; \omega) := \prod_{i} p(z_{i}; \omega_{i})$).
However, this could result in poor accuracy regarding the entropy estimation, because spatial neighborhood pixels are generally highly correlated and independent assumption causes over-estimation of the entropy. 
To solve this problem, \citet{li2017learning, rippel2017real, mentzer2018conditional} constructed a model of probability distribution for each quantized variable, given their neighborhoods.
This model is called~\textit{context model}.
However, via construction, using such a context model requires sequential encoding/decoding over $\comp$, thus the encoding/decoding speed significantly slows down when a computationally heavy model, such as PixelCNN~\cite{oord2016pixel}, is used for the context model.
To reduce the computational cost for the encoding/decoding, \citet{li2017learning, mentzer2018conditional} used learned \textit{importance masks} models on $\comp$, to adaptively skip the encoding/decoding by observing each estimated importance mask on each element of $\comp$.
\citet{mentzer2018conditional} achieved state-of-the-art performance using PixelCNN as the context model.
However, it still requires sequential encoding/decoding over $\comp$.

In contrast, our proposed model exhibits parallel encoding/decoding over a subset of $\comp$ 
assuming the conditional independence.
Our assumption of the conditional independence reflects the property of the natural image statistics, i.e.,
the spatial correlation between pixels decreases as the pixels are distant from each other.
The conditional independence allows to sample $\comp$ in a parallel manner, thus, our model can perform fast encoding/decoding with a GPU. 
\citet{reed2017parallel} demonstrated that such a parallel multi-scale density model achieves comparative performance with respect to the original PixelCNN, thus we can expect our proposed model to achieve both fast encode/decoding and accurate entropy coding.

Regarding the architecture of lossy autoencoder, to the best of our knowledge, all the existing studies used variants of the autoencoder, which exhibits a bottleneck at the deepest layer of the encoder. 
Although the proposed architecture possesses a similar structure 
as in the existing studies~\cite{ronneberger2015u, rippel2017real}
in the sense that the analyzer and the synthesizer exhibit symmetric structures, our model is specialized for lossy image compression where each feature map of the analyzer is quantized and stored 
to maintain the various resolutions of image features. This is not explored in existing studies where the quantization is applied to the deepest layer~\cite{balle2016end, theis2017lossy, Johnston2017improved, mentzer2018conditional} or applied after taking the sum of multiple features~\cite{rippel2017real}.
\citet{toderici2015variable, toderici2016full} proposed a different architecture compared with the standard lossy autoencoder we have described here, to realize variable compression rate regarding neural compression.
Their lossy compression model consists of recurrent neural network-based encoder and decoder. 
In each recurrence, the encoder considers the difference of each pixel between the original image and its reconstruction as an additional input to the originals, and
 the encoder and decoder are trained so as to minimize the distortion loss. 
Their proposed model, however, consumes significant computational time for encoding/decoding, compared with the standard lossy autoencoder model, owing to its nature of sequential recurrent computation.

\section{Experiments}\label{sec:setup}
\subsection{Experimental setup}
We conducted experiments to evaluate the image compression performance of our model regarding benchmark datasets.
We compared the performance with existing file formats, JPEG, WebP, BPG, and state-of-the-art neural compression methods~\cite{Johnston2017improved, rippel2017real}.

For the training dataset, we used Yahoo Flickr Creative Commons 100M~\cite{kalkowski2015real}, which has been used in the study regarding the current state of the art method~\cite{rippel2017real}.
The original dataset consists of 100-million images.
We selected portions of images whose both vertical and horizontal resolution were greater than or equal to $1{,}024 \times 1{,}024$. 
We used 95{,}205 selected images for training and 1{,}000 selected images for validation.
For pre-processing of the lossy autoencoder training, we resized the images into those whose short sides were 512.
Subsequently, we cropped the resized images to $512 \times 512$.
For the lossless coder training, we performed the same pre-processing as for the lossy autoencoder, except that our resizing and cropping size was $1{,}024 \times 1{,}024$.
We used (negative) MS-SSIM~\cite{wang2003multiscale} for the distortion loss~\eqref{eq:lossy_loss}, which is observed to exhibit high correlation with human subjective evaluation. 
It is commonly used to evaluate the quality of the image compression. 
Please refer to the appendix for further details of 
the experiments.

\subsection{Comparison of the compression performances}\label{sec:results}
Fig.~\ref{fig:eval_kodak}\footnote{Regarding the RD curves of \citet{rippel2017real}, we carefully traced the RD curve from the figure of their paper, because we could not obtain the exact values at each point. 
As for the RD curve of \citet{Johnston2017improved}, we used the points provided by the authors via personal communication. The exact values of bpp and MS-SSIM on the RD curve of the proposed method are shown in Table 1 in appendix.} shows the RD curves with different compression methods on the Kodak dataset.
Our proposed model achieved superior performance to 
the existing file formats, 
JPEG\footnote{\url{http://www.ijg.org/}},
WebP\footnote{\url{https://developers.google.com/speed/webp/}},
and BPG\footnote{\url{https://bellard.org/bpg/}}.

Moreover, the proposed model achieved better performance than 
nearly all the existing neural compression methods~\cite{toderici2016full, balle2016end}.
It demonstrated comparable performance with respect to recent CNN-based compression~\cite{rippel2017real}.
When we compare the RD-curve carefully with \citet{rippel2017real}, 
it seems that our method is advantageous in case of wide range of low bit-rates.
Refer to appendix for certain reconstructed images on the Kodak dataset.

\begin{figure}[htbp]
    \centering
    \includegraphics[width=0.85\linewidth]{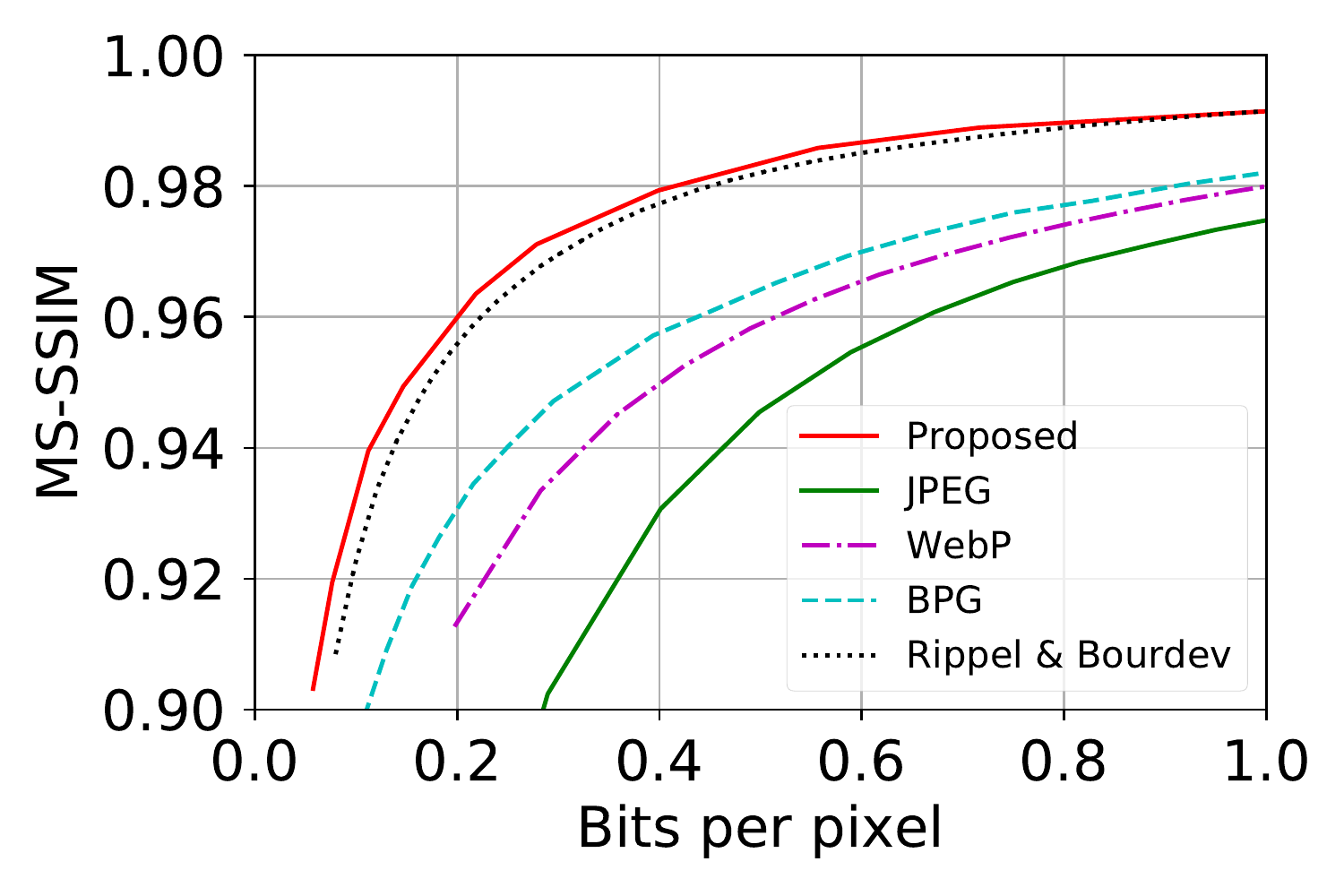}
    \caption{Rate-distortion trade-off curves with different methods on the RAISE-1k dataset.}
    \label{fig:eval_raise_crop}
\end{figure}

Fig.~\ref{fig:eval_raise_crop} shows the RD curves on the RAISE-1k dataset.
Our model also achieves superior performance over the other methods.

To evaluate the encoding and decoding speeds of our model, 
we compare the computation time for encoding and decoding against JPEG, WebP, and BPG. Unfortunately, we cannot implement the state-of-the-art DNN-based study~\cite{rippel2017real} because the detail of the architecture and lossy compression procedures are not written in their paper.

As for the computational resources, our proposed model operated on a single GPU and a single CPU process, whereas JPEG, WebP, and BPG used a single CPU process. 
We used PNG file format as original images because it is a widely accepted file format and publicly available BPG codec, libbpg, only accepts either JPEG or PNG file format as the input. 
To achieve a fair comparison, we measured the computation time including the decoding of PNG-format image into Python array, and encoding process from Python array into the binary representations (encoding time), and the time required for decoding from the binary representations into the PNG-format image (decoding time).
Note that the transformation between a PNG-format image and a GPU array was not optimized with respect to the computation time; we did not transform each PNG-format image into a cuda array on GPU directly\footnote{We first transformed the PNG-format image into a numpy array on CPU with the python library `Pillow', then transferred it to a cuda array on GPU.}. 
The direct transformation, which would reduce the computation time of our model, is reserved for future study.

\begin{figure}
    \centering
    \includegraphics[width=1.0\linewidth]{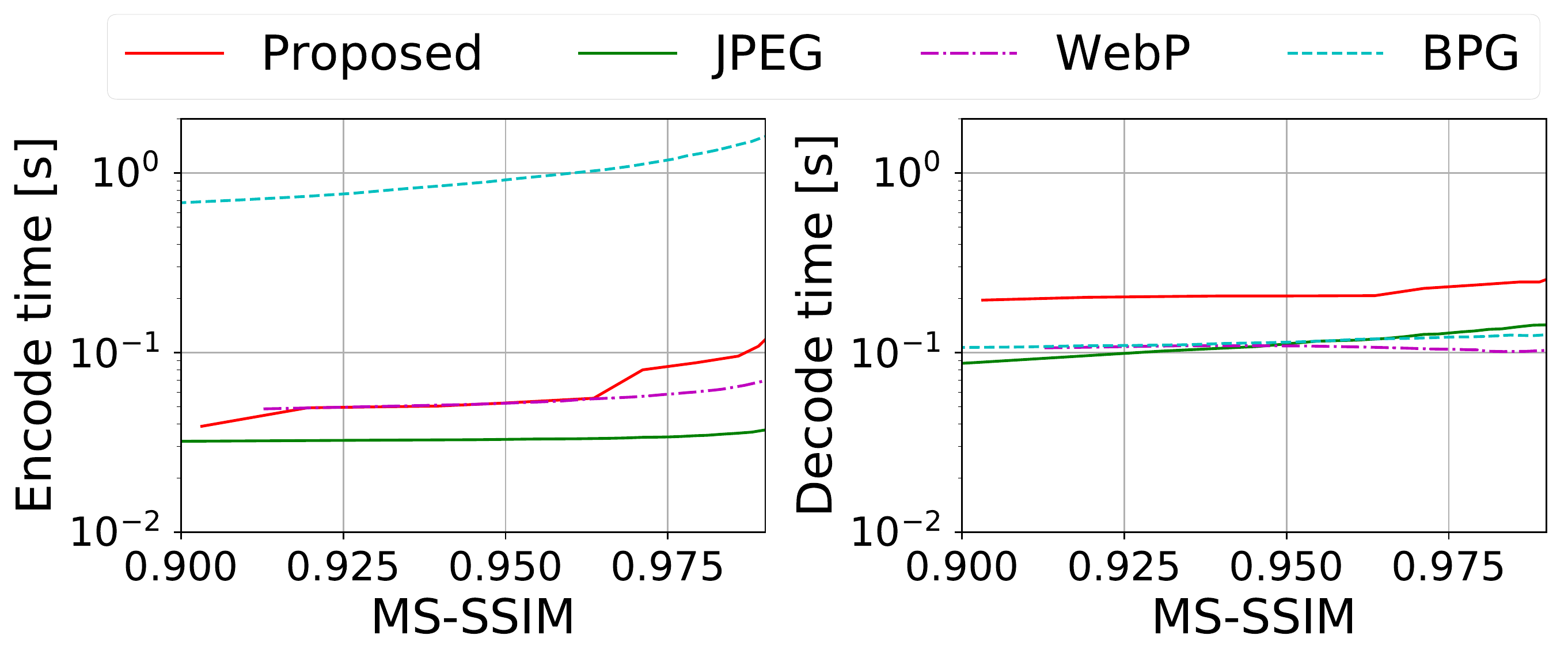}

    \caption{
    Computational time vs. MS-SSIM plots on the RAISE-1k dataset.
    The left and right panel indicate the results of the encoding time and decoding time, respectively.}
    \label{fig:time_raise_crop}
\end{figure}

Fig.~\ref{fig:time_raise_crop} shows the computational times for encoding and decoding. 
For encoding, our proposed method takes approximately 0.1 s or less than 0.1 s for the region where the MS-SSIM takes less than 0.98.
It is significantly faster than BPG, but inferior to JPEG and WebP.
Note that a reconstructed image with an MS-SSIM of 0.96
is usually a high-fidelity image, which is difficult to distinguish from the original image by human eyes.
Two examples of our reconstructed image are shown in Fig.~\ref{fig:comp_reconst}.
The MS-SSIM of the top and bottom reconstructed image are 0.961 and 0.964, respectively.
The decoding time of our proposed method is approximately 0.2 s for the range of MS-SSIM between 0.9 to 0.99.
It is nearly two times slower than the other file formats.

\subsection{Analysis by ablation studies}
We conducted ablation experiments to indicate the efficacy of each of our proposed modules: multi-scale lossy autoencoder and parallel multi-scale lossless coder.

\subsubsection{Parallel multi-scale lossless coder vs. independent lossless coder}
\begin{figure}[htbp]
    \centering
    \includegraphics[width=0.8\linewidth]{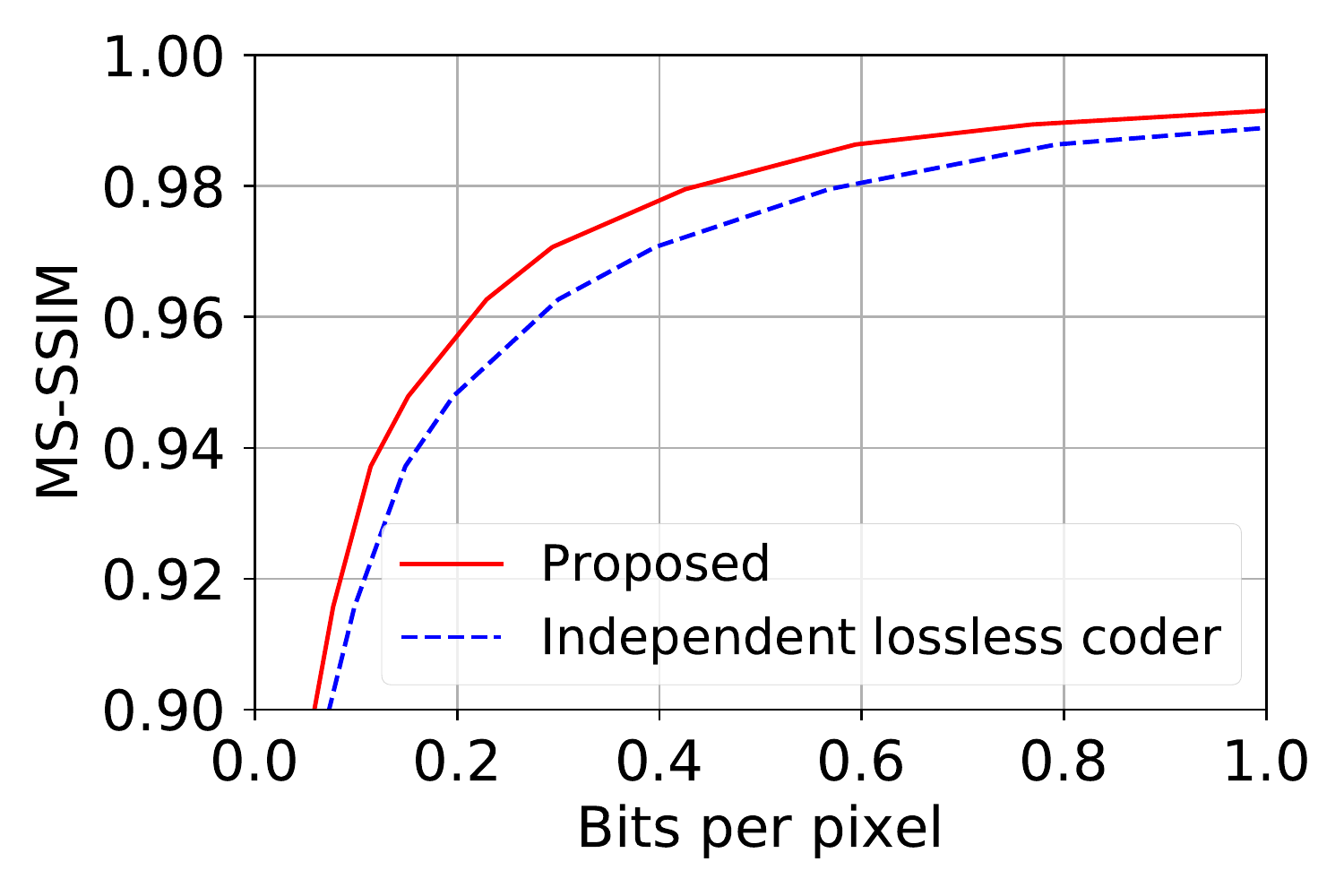}
    \caption{Entropy coding by proposed parallel multi-scale lossless coder vs. independent lossless coder.}
    \label{fig:compare_lossless}
\end{figure}
Here, we discuss the efficacy of our lossless coder against an \textit{independent lossless coder} 
that conducts lossless coding of each element of $\comp$ by observing 
the histograms of each channel of $\comp$.
As for the lossy autoencoder model, we used architecture identical to the proposed model.
Fig.~\ref{fig:compare_lossless} shows the RD-curves with respect to our 
lossless coder and independent lossless coder~\cite{balle2016end, theis2017lossy}.
We can observe that our lossless coder achieved significantly better performance than the independent lossless coder. 
Note that, evidently, the independent lossless coder can encode/decode faster compared with our lossless coder.
\subsubsection{Multi-scale vs. single-scale autoencoder}
\begin{figure}[htbp]
    \centering
    \includegraphics[width=0.8\linewidth]{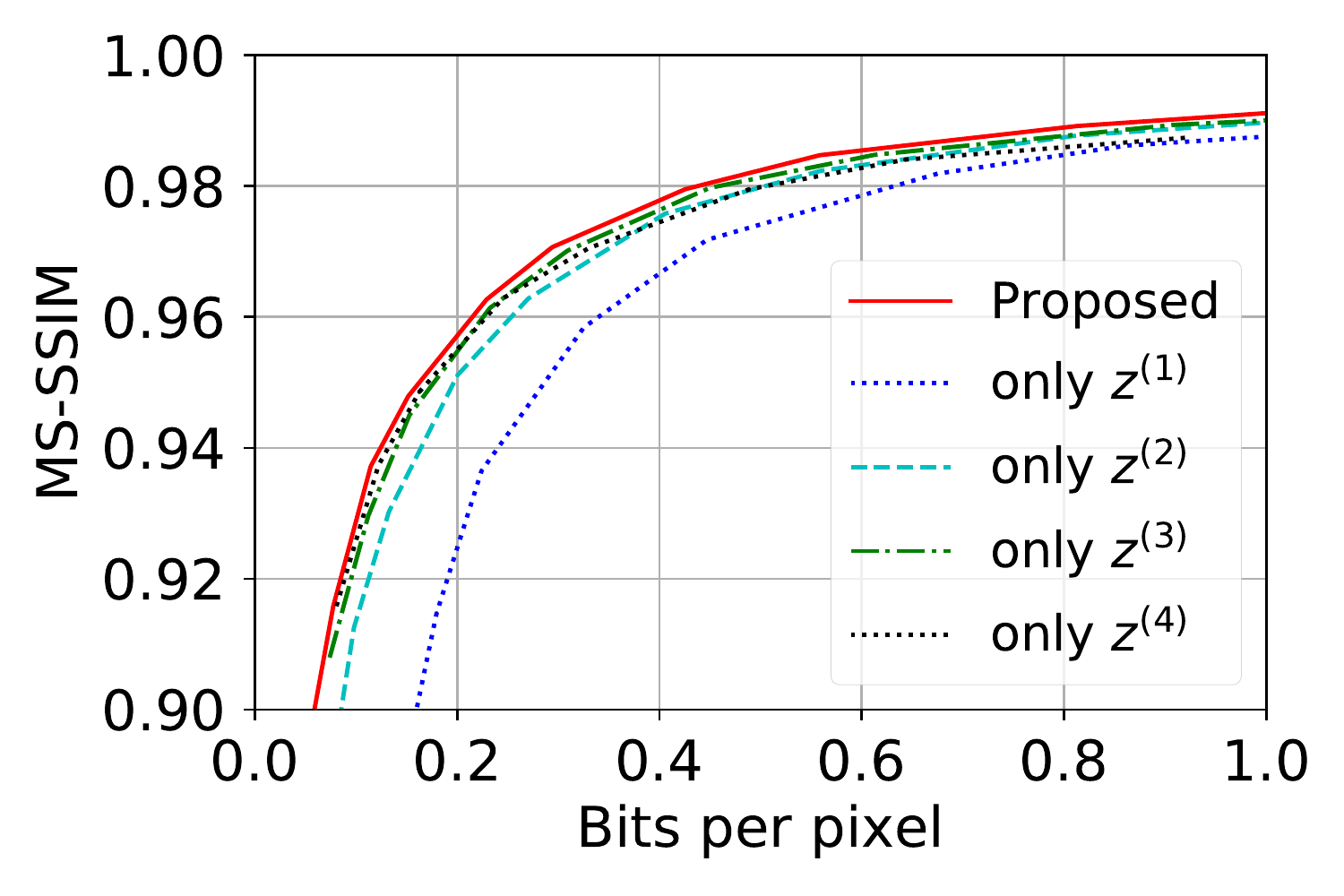}
    \caption{Multi-scale vs. single-scale autoencoder. 
        RD-curves of our proposed model (multi-scale autoencoder) are compared with that of 
        single-scale autoencoders. Each single-scale autoencoder possesses architecture 
    identical to the proposed model, but the quantization is applied to the units only at specific depth of the layer.
   ``only $z^{(i)}$" in the legend refers to the single-scale autoencoder 
    that is quantized only at the $i$-th layer units.   
    }
    \label{fig:compare_resolution}
\end{figure}
Fig.~\ref{fig:compare_resolution} shows the RD-curves on multi-scale (i.e., proposed) and single-scale autoencoders.
Each single-scale autoencoder possess architecture 
identical to the proposed model, except it only possess only one connection at specific depth of the layer.
As for the structure of each single-scale autoencoder, we used the architecture identical to the proposed model, except that the quantization is applied at a specific depth of the layer.
The deeper layers are not used for both encoding and decoding.
The training was separately conducted to achieve the best performance for each single-scale autoencoder.
In this experiment, the quantization level is always fixed at 7.
As can be observed from the figure, our model certainly achieved better performance than 
other single-resolution autoencoder models at any bitrate,
although, the gains are not so large. 
This may be because we did not conduct the joint optimization 
of the parameters of the lossy autoencoder and the lossless coder. 
The separate optimization may hinder exploitation of the benefits of multi-resolution features, 
because the entropy of the feature map would be different at each resolution, 
and the separate optimization makes it difficult to exploit the property.

Fig.~\ref{fig:comp_image} shows the quantized feature map at each layer, obtained via our lossy autoencoder. 
We observe that each quantized feature map extracts different scale of features. 
For example, the quantized feature maps at the high-resolution layer (top panels) encode the features in the high-frequency domain, including edges, whereas the quantized feature map at the low-resolution layer encodes the features in the low-frequency domain, including background colors and surfaces.

\section{Conclusion}
In this study, we propose a novel CNN-based lossy image compression algorithm.
Our model consists of two networks: \textit{multi-scale lossy autoencoder} and \textit{parallel multi-scale lossless coder}.
The multi-scale lossy autoencoder extracts multi-scale features and encodes them. 
We successfully obtained different features of images. Local and fine information, such as edges, were extracted at the relatively shallow layer, and global and coarse information, such as textures, were extracted at the deeper layer. 
We confirmed that this architecture certainly improves the RD-curve at any bitrate. 
The parallel multi-scale lossless coder encodes the discretized feature map into compressed binary codes, and decodes the compressed binary codes into the discretized feature map in a lossless manner. 
Assuming the conditional independence and the parallel multi-scale pixelCNN ~\cite{reed2017parallel}, we encoded and decoded the discretized feature map in a partially parallel manner, making the encoding/decoding times significantly fast without losing much quality. 
Our experiments with the Kodak and RAISE-1k datasets indicated that our proposed method achieved state-of-the-art performance with reasonably fast encoding/decoding times. 
We believe our model makes the CNN-based lossy image coder step towards the practical uses that require high image compression quality and fast encoding/decoding time.

\bibliography{bibliography}
\bibliographystyle{icml2018}

\newpage
\appendix

\section{Lossless encoding of multi-scale features}
\begin{figure}[htbp]
    \centering
    \includegraphics[width=0.8\linewidth]{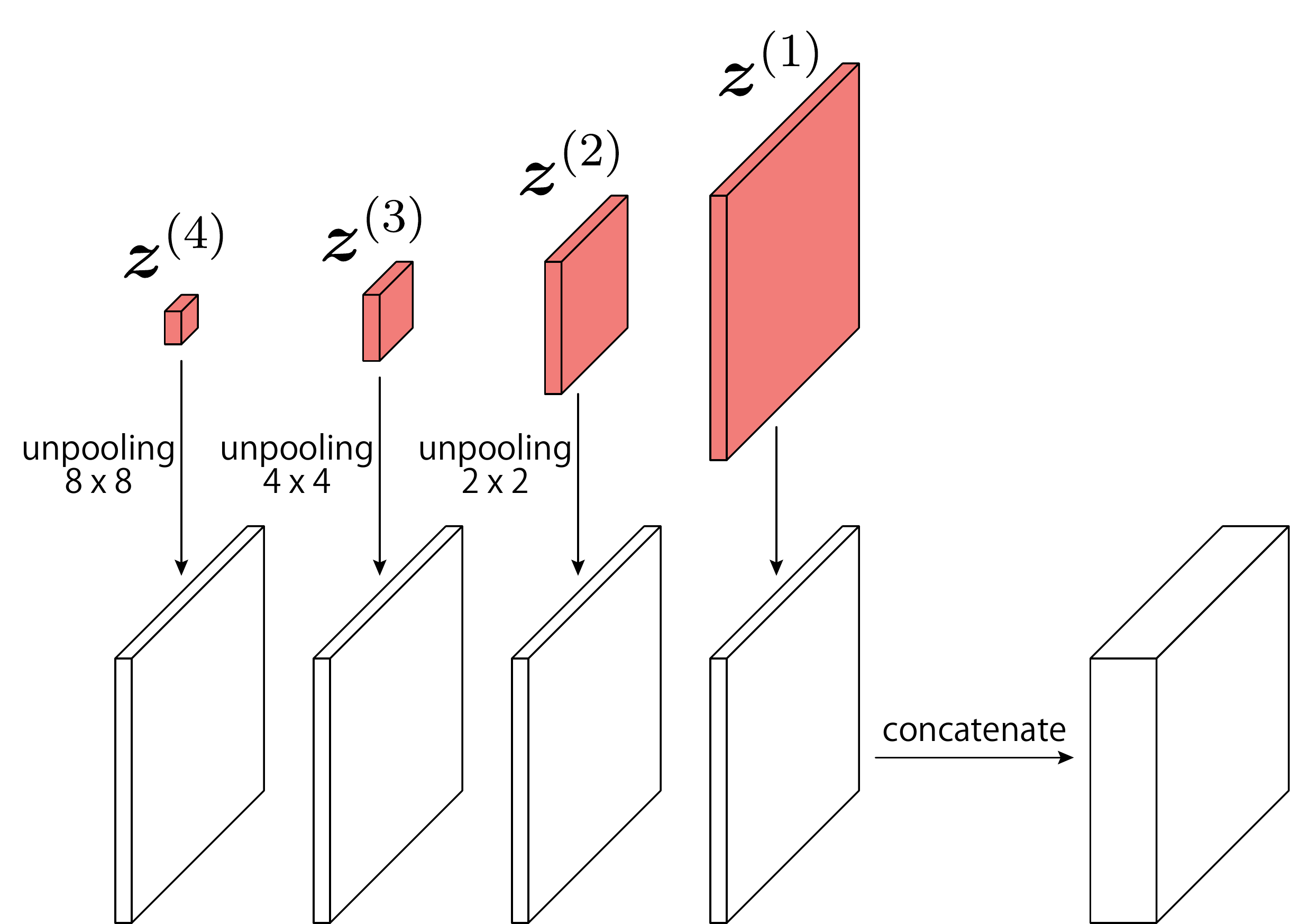}
    \caption{
Integration of multi-scale feature map \mnote{multi-scale feature map"s"?}. The feature map is upsampled 
to maintain the same spatial resolution as the densest feature map obtained via unpooling. 
The upsampled feature map is accumulated \mnote{multi-scale feature maps are?} and all channels are concatenated.
    }
    \label{fig:Multiscale Encoding}
\end{figure}
In our proposed architecture, several feature maps, 
each having different resolutions, are quantized. 
To encode different scaled feature maps, we upsample them, 
i.e., \mnote{weが抜けてる？} copy their values to the interpolated adjacent units (i.e., unpooling) 
so that all upsampled feature maps possess the same resolution 
as the densest feature map. 
Subsequently, the channels are accumulated with respect to the corresponding unit, as shown in Fig.~\ref{fig:Multiscale Encoding}.
Next, the dark-gray units in Fig.~\ref{fig:lossless_unit} integrate the information of all the channels provided by all resolution feature maps.
Similarly, the red units contain the information of all channels. 
Certain channels are shared between dark-gray units and red units because of unpooling.
Thus, the non-trivial channels of red units, which are not shared with dark-gray units, 
are estimated for the training of the conditional distribution.

\section{Detail of the experiment}
\subsection{Evaluation method}
Regarding the distortion loss (1),
we used MS-SSIM~\cite{wang2003multiscale}, which is demonstrated to exhibit high correlation with human subjective evaluation. 
It is commonly used to evaluate the quality of the image compression. 
MS-SSIM was originally designed for single channel images. 
To use MS-SSIM for RGB images, we calculated MS-SSIM with respect to each RGB channel and reported the average of the values. 
We evaluated MS-SSIMs at multiple compression rates to draw 
the rate-distortion trade off curve (RD-curve).
For this evaluation, we used Kodak and RAISE-1k~\cite{Dang2015raise} dataset. The Kodak dataset consists of 24 natural images of size $512 \times 768$, and the RAISE-1k dataset consists of 1K natural images of various image sizes.
All of the original RAISE-1k image sizes are exceedingly large, 
which prevents the evaluation of the MS-SSIM scores in a reasonable computation time.
Thus, we resized and center-cropped each original image as pre-processing.
Subsequently, we evaluated the $512 \times 768$ and $768 \times 512$ pre-processed images.

We calculated bits-per-pixel (bpp) for each method 
via observing the actual file size of the binary variables. 
Although the compressed file may possess included meta-information, we did not remove it, assuming the size of the meta-information was negligible, compared to the size of the main body of the binary file.

\subsection{Architecture}
\begin{figure*}[htbp]
    \centering
    \includegraphics[width=0.8\linewidth]{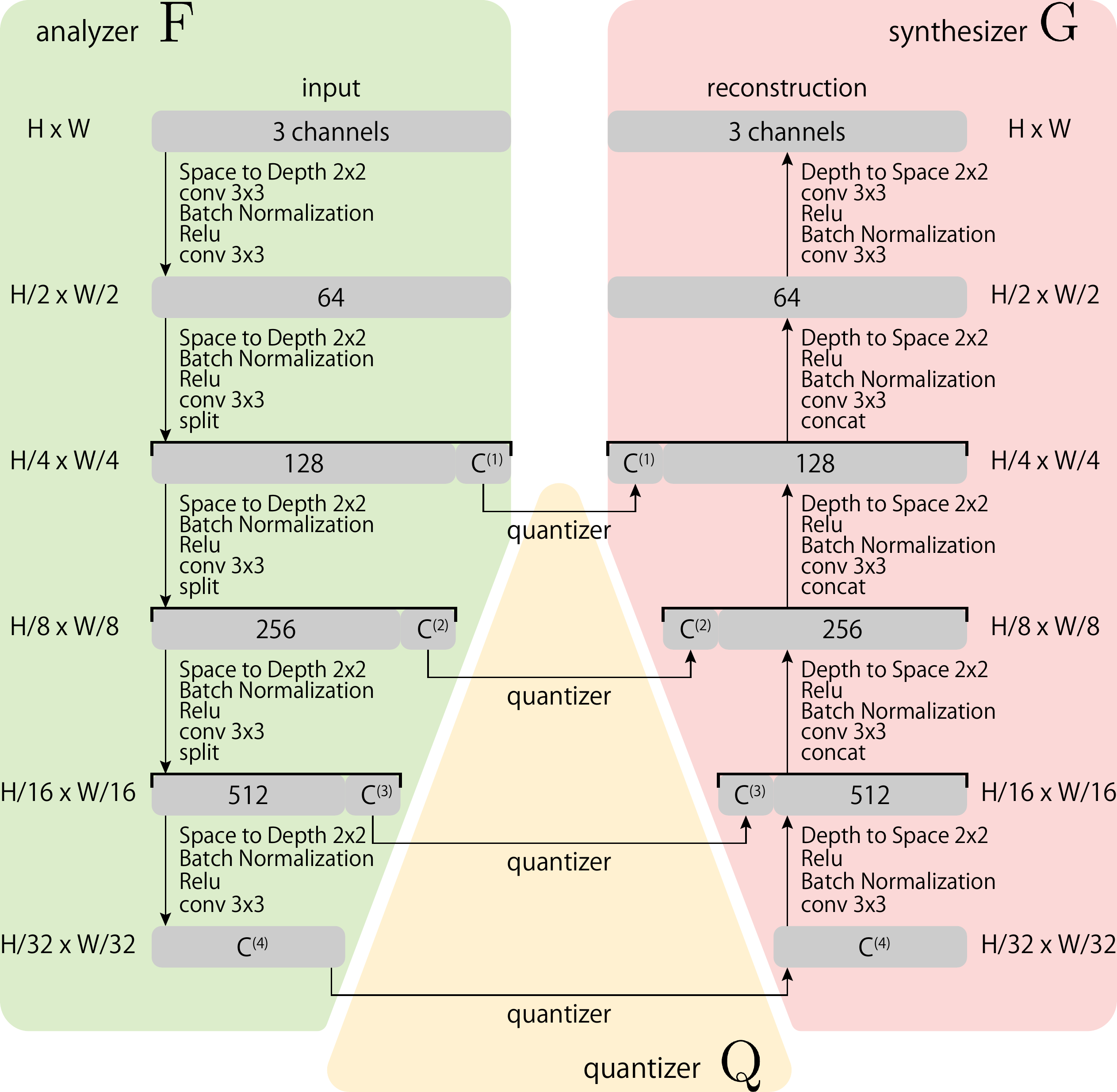}
    \caption{Architecture of lossy autoencoder}
    \label{fig:lossy_network_detail}
\end{figure*}

The proposed lossy autoencoder was composed of $6$ convolutional layers and we extracted the quantized variables at four deepest layers as shown in Fig.~\ref{fig:lossy_network_detail}.
The number of feature maps (i.e., the number of channels $C^{(i)}$) 
of each $\comp^{(i)}$ varied at each compression rate.
The number is optimized as follows.
First, we set the number of channels at the deepest layer as 32.
Subsequently, the number of channels at the shallower layer is set so that 
it gradually decreases as the layer becomes shallow. 
Because the number of pixels (i.e., $H^{(i)} \times W^{(i)}$) 
is four times larger than that of the adjacent higher layer, 
we search around $C^{(i)} \approx C^{(i+1)}/4$ 
to maintain similar amount of information (i.e., $C^{(i)} \times H^{(i)} \times W^{(i)} \approx C^{(i+1)} \times H^{(i+1)} \times W^{(i+1)} $) at each layer.
For accuracy, $C^{(i)}$ were optimized from $C^{(i)} \in [0, C^{(i+1)}]$.
We also optimized the quantization level.
The quantization level was selected either $7$ or $13$ irrelevant to the layer.
We select $7$ at low-bit-rate coding while we select $13$ at high-bit-rate coding.
The quantization level and the number of channels used to reproduce the RD curve of the proposed method in Figs. 1 and 8 are summarized in Table 1.

As for the parallel multi-scale lossless coder, we used a CNN with $4$ convolution layers, which we call a block, to approximate each $p(v^{(k)}_i | \comps^{(1)}, \cdots, \comps^{(k-1)})$ $(k=2,\cdots,K)$ in Eq.~\eqref{eq:subset}. 
We select the number of blocks $K$ from $\{8,10,12\}$ depending on the bitrate. 
Note that a pair of blocks corresponds to the conditional lossless coder whose coding process is depicted in Fig.~\ref{fig:lossless_unit}(d).

At the test phase, we tested two architectures; the one consists of an identical architecture with that used in the training, and the other consists of a smaller architecture where the last two blocks of the architecture used in the training are not used, i.e., the number of blocks are $K-2$ where $K$ is the number of blocks used during the training.
Hence $z^{(1)}$ at the test phase is denser than that of $z^{(1)}$ at the training phase, and the denser $z^{(1)}$ at the test phase is encoded as $\save$ using the independent lossless coder. 
The latter architecture could deteriorate the compression rate because it does not consider the correlation among the denser pixels and the testing condition is different from the training condition. 
However, we observed that the latter model does not decrease the compression rate considerably while it exhibits faster encoding/decoding.
Therefore, we present the experimental results of the latter model.

\subsection{Training}
We trained each model with 100{,}000 updates using mini-batch stochastic gradient decent. 
The batchsize was $24$ for the lossy autoencoder and $6$ for the lossless coder.
We used Adam optimizer~\cite{kingma2014adam} with the hyper-parameters,
 $\alpha = 0.001$, $\beta_1 = 0.9$, and $\beta_2 = 0.999$.
We also applied linear decay for the learning rate, 
$\alpha$, after 75{,}000 iterations so that the rate would be 0 at the end.

\begin{table*}[hbp]
\centering
\begin{tabular}{cc|cc|ccccc}
\hline
\multicolumn{2}{c|}{Kodak} & \multicolumn{2}{|c|}{RAISE-1K} &&&&& \\
bpp & MS-SSIM & bpp & MS-SSIM & quantization level & $C^{(1)}$ & $C^{(2)}$ & $C^{(3)}$ & $C^{(4)}$ \\
\hline
\hline
0.0582 & 0.8994 & 0.0576 & 0.9032 &  7 & 0 &  0 &  4 & 32 \\
0.0775 & 0.9157 & 0.0766 & 0.9195 &  7 & 0 &  1 &  4 & 32 \\
0.1146 & 0.9372 & 0.1123 & 0.9396 &  7 & 0 &  2 &  8 & 32 \\
0.1517 & 0.9479 & 0.1468 & 0.9494 &  7 & 0 &  3 & 12 & 32 \\
0.2292 & 0.9627 & 0.2186 & 0.9636 &  7 & 0 &  5 & 20 & 32 \\
0.2942 & 0.9707 & 0.2786 & 0.9711 &  7 & 1 &  4 & 24 & 32 \\
0.4256 & 0.9795 & 0.3987 & 0.9793 &  7 & 2 &  8 & 24 & 32 \\
0.5939 & 0.9864 & 0.5567 & 0.9858 & 13 & 2 &  8 & 24 & 32 \\
0.7680 & 0.9894 & 0.7166 & 0.9890 & 13 & 3 & 12 & 24 & 32 \\
1.0925 & 0.9924 & 1.0289 & 0.9917 & 13 & 5 & 20 & 24 & 32 \\
\hline
\end{tabular}
\caption{Bits-per-pixel (bpp) and MS-SSIM on RD curve of the proposed method in Figs. 1 and 8. The corresponding hyperparameters (quantization level and number of channels) of the architecture used for the compression are also shown in the right column.}
\end{table*}

\section{Examples of reconstruction images}
\begin{figure*}[t]
    \centering
    \includegraphics[width=0.75\linewidth]{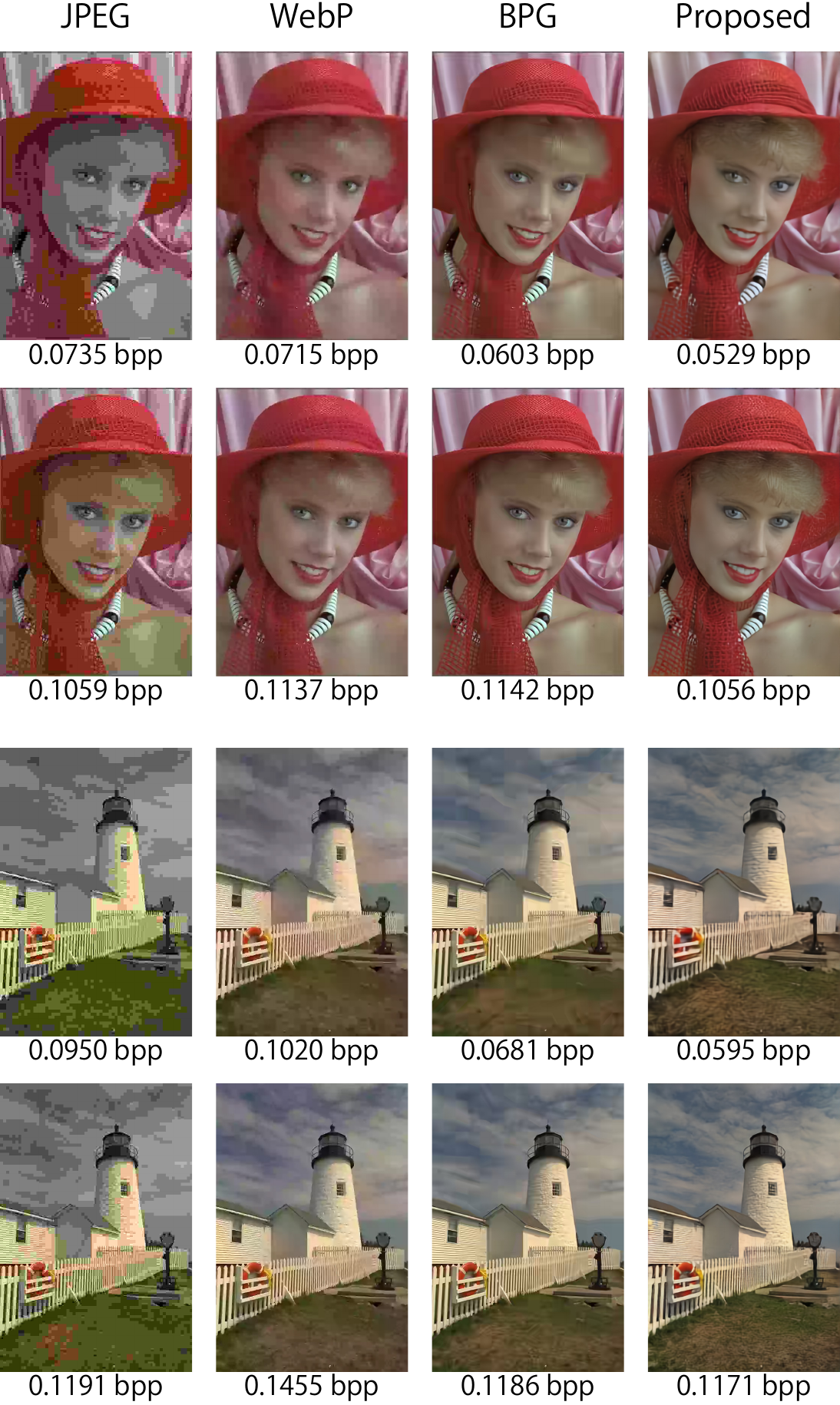}
    \caption{Comparison of reconstruction images of size 768 $\times$ 512. The number at the bottom of each panel indicates bit-per-pixel (bpp) of each reconstructed image.
    }
    \label{fig:rec1}
\end{figure*}

\begin{figure*}[t]
    \centering
    \includegraphics[width=0.99\linewidth]{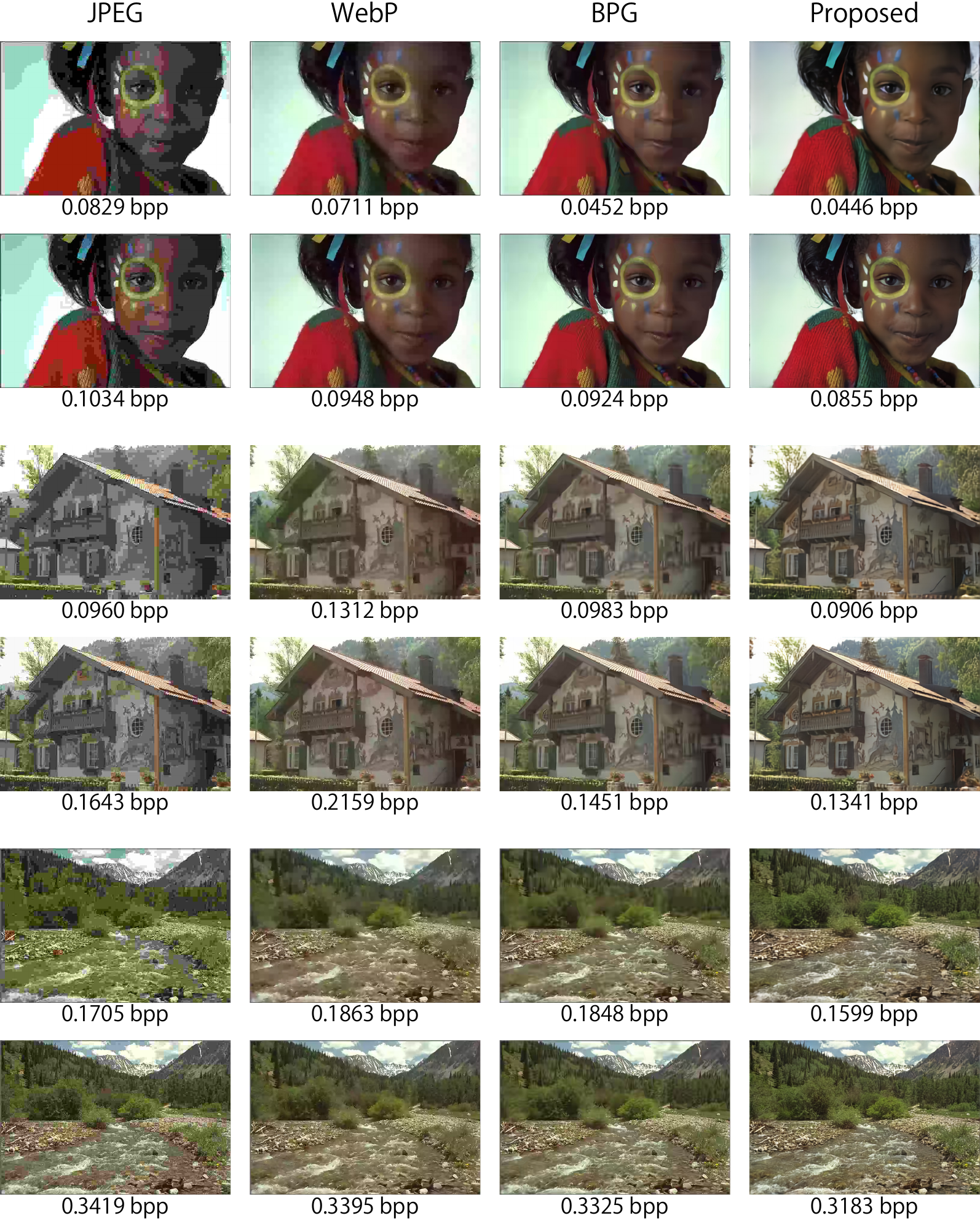}
    \caption{Comparison of reconstruction images of size 512 $\times$ 768. The number at the bottom of each panel indicates bit-per-pixel (bpp) of each reconstruction}
    \label{fig:rec2}
\end{figure*}

Figs.~\ref{fig:rec1} and~\ref{fig:rec2} show the reconstructed images obtained with different compression methods.
From left to right, we demonstrate the reconstructed images with JPEG, WebP, BPG, and our proposed method.
Owing to the size limitation of the supplemental file, 
we compress each reconstructed images via JPEG with the highest quality.

\end{document}